\documentclass[sigconf]{acmart}
\usepackage{graphicx}
\usepackage{cuted}
\usepackage{hyperref}
\usepackage{xcolor}

\AtBeginDocument{%
  \providecommand\BibTeX{{%
    \normalfont B\kern-0.5em{\scshape i\kern-0.25em b}\kern-0.8em\TeX}}}

\copyrightyear{2024}
\acmYear{2024}
\setcopyright{acmlicensed}\acmConference[SIGIR '24]{Proceedings of the 47th International ACM SIGIR Conference on Research and Development in Information Retrieval}{July 14--18, 2024}{Washington, DC, USA}
\acmBooktitle{Proceedings of the 47th International ACM SIGIR Conference on Research and Development in Information Retrieval (SIGIR '24), July 14--18, 2024, Washington, DC, USA}
\acmDOI{10.1145/3626772.3657680}
\acmISBN{979-8-4007-0431-4/24/07}

\begin{document}

\title{\textsc{ResumeFlow}: An LLM-facilitated Pipeline for Personalized Resume Generation and Refinement}

\author{Saurabh Bhausaheb Zinjad}
 \affiliation{%
  \institution{Arizona State University}
  \city{Tempe}
  \state{Arizona}
  \country{USA}
}
\email{szinjad@asu.edu}


\author{Amrita Bhattacharjee}
\affiliation{%
  \institution{Arizona State University}
  \city{Tempe}
  \state{Arizona}
  \country{USA}}
\email{abhatt43@asu.edu}

\author{Amey Bhilegaonkar}
\affiliation{%
  \institution{Arizona State University}
  \city{Tempe}
  \state{Arizona}
  \country{USA}}
\email{abhilega@asu.edu}

\author{Huan Liu}
\affiliation{%
  \institution{Arizona State University}
  \city{Tempe}
  \state{Arizona}
  \country{USA}
}
\email{huanliu@asu.edu}


\begin{abstract}
Crafting the ideal, job-specific resume is a challenging task for many job applicants, especially for early-career applicants. While it is highly recommended that applicants tailor their resume to the specific role they are applying for, manually tailoring resumes to job descriptions and role-specific requirements is often (1) extremely time-consuming, and (2) prone to human errors. Furthermore, performing such a tailoring step at scale while applying to several roles may result in a lack of quality of the edited resumes. To tackle this problem, in this demo paper, we propose \textsc{ResumeFlow}: a Large Language Model (LLM) aided tool that enables an end user to simply provide their detailed resume and the desired job posting, and obtain a personalized resume specifically tailored to that specific job posting in the matter of a few seconds. Our proposed pipeline leverages the language understanding and information extraction capabilities of state-of-the-art LLMs such as OpenAI's GPT-4 and Google's Gemini, in order to (1) extract details from a job description, (2) extract role-specific details from the user-provided resume, and then (3) use these to refine and generate a role-specific resume for the user. Our easy-to-use tool leverages the user-chosen LLM in a completely off-the-shelf manner, thus requiring no fine-tuning. We demonstrate the effectiveness of our tool via a \textcolor{blue}{\href{https://www.youtube.com/watch?v=Agl7ugyu1N4}{video demo}} and propose novel task-specific evaluation metrics to control for alignment and hallucination. Our tool is available at \textcolor{blue}{\textbf{\href{https://job-aligned-resume.streamlit.app}{\textsc{ResumeFlow}}}}.

\end{abstract}





\maketitle

\section{Introduction}

One of the most important steps for a candidate on the job market to focus on is crafting a good quality, impactful resume. Condensing years of education, projects and unique experiences into a short 1 or 2 page document while still highlighting the uniqueness and individuality of the candidate is a quite challenging task. Given the increasing number of candidates applying to jobs, employers often rely upon automated applicant tracking systems (ATS) and resume filtering systems that filter out candidates based on the degree of match to a job opening. In order to deal with these automated systems, candidates now need to spend time and effort crafting and tailoring their resume to the specific job posting, such as retaining only relevant experience and skills, highlighting job-specific projects, highlighting achievements, etc. This is a challenging and extremely time-consuming step and many candidates struggle with identifying and using impactful keywords, keeping the resume concise, etc~\cite{Tom_2022,Res_1}. While there are some automated tools for analyzing and retrieving information from candidate resumes, there is currently no automated solution via which a job applicant can tailor their resume to some specific job they are interested in. To solve this issue, in this work we propose a tool called \textsc{ResumeFlow} that enables the user to simply use their general-purpose resume and tailor it to a specific job that the user is interested in, thereby potentially saving hours of effort for the applicant. 

Recent advancements in language modeling and particularly in the direction of instruction-tuned Large Language Models (LLMs) have demonstrated the vast capabilities of such models. Alongside improved performance on standard natural language tasks~\cite{chang2023survey}, LLMs have also been shown to have impressive performance in many challenging use-cases~\cite{bubeck2023sparks,bhattacharjee2023llms}. Inspired by the impressive instruction following capabilities~\cite{zhou2023instruction} of recent LLMs, such as OpenAI's GPT-3.5, GPT-4~\cite{achiam2023gpt} and Google Deepmind's Gemini~\cite{team2023gemini}, we aim to leverage the textual understanding, instruction following, and generation capabilities of these recent LLMs to facilitate the task of resume tailoring. Therefore, we design and develop \textsc{ResumeFlow} as an end-to-end pipeline that leverages the power of Large Language Models for retrieval, document processing, and text generation, especially in the context of user resumes. Our tool is available at \textcolor{blue}{\url{https://job-aligned-resume.streamlit.app}} and a demo video is also available at \textcolor{blue}{\href{https://www.youtube.com/watch?v=Agl7ugyu1N4}{this YouTube link}}.

\begin{figure*}[!ht]
    \centering
    \includegraphics[width=7in]{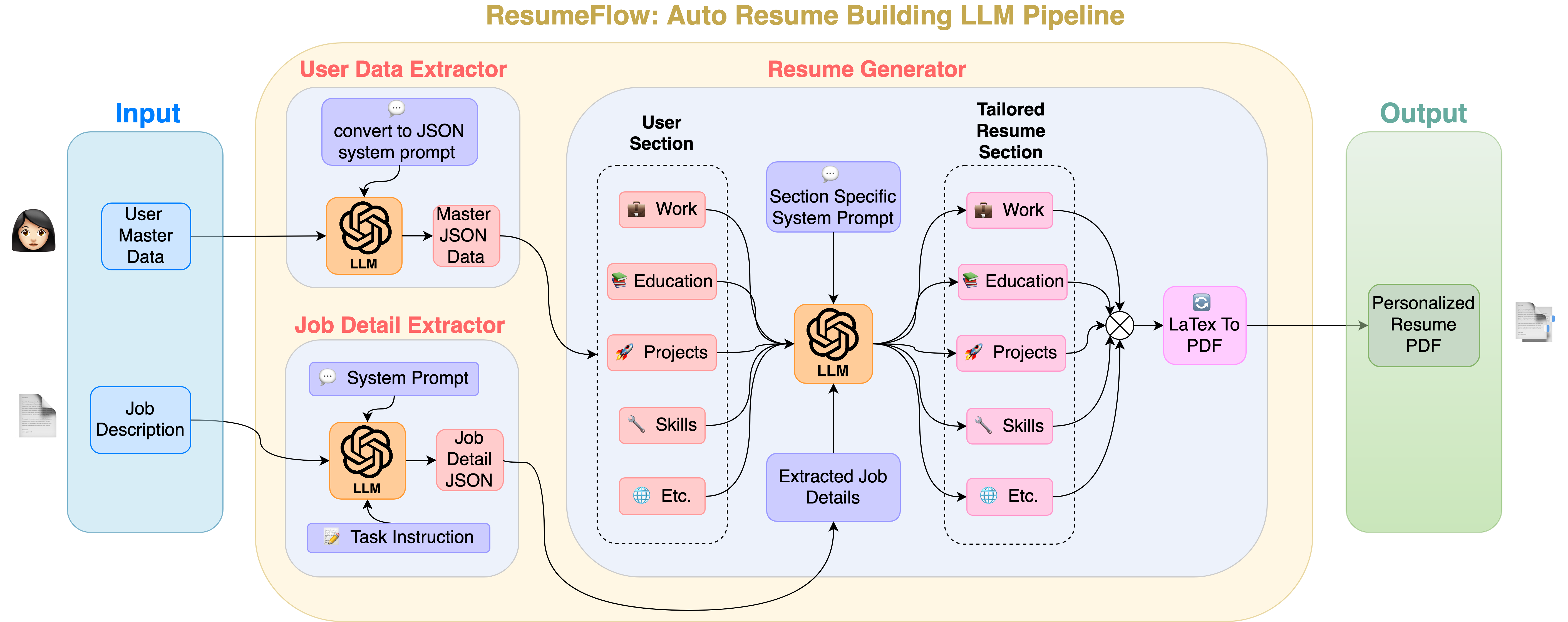}
    \caption{Our overall \textsc{ResumeFlow} pipeline for LLM-facilitated generation of job-aligned resumes.}
    \label{fig:pipeline}
\end{figure*}

\section{Related Work}

With the evolution of various natural language and text mining methods, there have been interesting applications of new methods in tasks such as resume analysis and classification, information retrieval from resumes, etc.  Several of these efforts try to parse out and extract information ~\cite{li2021method,tallapragada2023improved}, predict skills from resumes~\cite{jiechieu2021skills}, perform job matching~\cite{yi2007matching}, etc. Authors in ~\cite{chen2018two} propose a two-step approach for information extraction from resumes, by identifying the blocks of information first. However, to the best of our knowledge, there has been no work on automatically generating job-aligned resumes. In the direction of using LLMs as instruction-following assistants, there has been recent work in the direction of using LLMs for information retrieval~\cite{wei2023zero,zhu2023large}, along with interesting explorations in the field of education~\cite{lo2023impact,adeshola2023opportunities}, text analysis and research~\cite{pal2023domain,meyer2023chatgpt}, etc. Inspired by these works and the overall potential of recent state-of-the-art LLMs, we propose to leverage LLMs in our \textsc{ResumeFlow} pipeline.
A recent work that is closest to ours is ~\cite{skondras2023efficient} where the authors generate structured and unstructured resumes using ChatGPT. However, they simply use it as a data generation/augmentation step for a downstream classification task, and there is no aspect of tailoring the resumes to specific jobs.

\section{\textsc{ResumeFlow}: How It Works}

In this section, we describe how our \textsc{ResumeFlow} tool works in the backend and describe each of the components in detail.
Figure \ref{fig:pipeline} presents an overview of \textsc{ResumeFlow}, our proposed LLM-facilitated pipeline for personalized resume generation. The system comprises three core components:
\begin{itemize}
    \item \textbf{User Data Extractor:} This module transforms a user-provided baseline resume, in the form of a PDF document, into a structured JSON format suitable for subsequent processing.
    \item \textbf{Job Details Extractor}: This module analyzes the job description, provided as a string by simply copy-pasting the job posting, to extract keywords, requirements, etc., in the form of a JSON, and will be used in the next step for informing the content and tailoring of the generated resume.
    \item \textbf{Resume Generator}: This component is where the bulk of the processing takes place. Leveraging the structured user data and extracted job requirements, this module generates a personalized resume.
\end{itemize}

The following sections will introduce details of the pipeline.

\subsection{User Data Extractor}

Our tool allows the user to input their (structured or unstructured) resume as a PDF, along with the job description of the job they are applying to. The user also has the option to choose which LLM to be used for the entire pipeline. In this component, we first convert the resume PDF to a text string and then use the user-chosen LLM as an information extraction model to extract key sections and important information from the user's resume text. To do this, we simply use the chosen LLM in an off-the-shelf manner without additional training, by using carefully structured prompts that facilitate information extraction. To perform this step, we use a prompt such as (entire prompt is truncated for brevity):
 ``You are a software developer working on a resume parsing application. 
Your task is to design a system that takes a resume in text format as input and converts it into a structured JSON format."

This allows us to have a structured dictionary of the user's information as extracted by the LLM. We refer to this dictionary as \textbf{UserData} in the following sections, for ease of understanding.

\subsection{Job Details Extractor}

The ``Job Details Extractor" accepts job descriptions in text format: the user is instructed to copy-paste the job description from their desired job listing into the textbox on our tool. 
This component in our pipeline aims to extract the key points from the job description such as: job title, required and preferred qualifications, etc. In order to do this via the LLM, we use two carefully crafted prompts: the system prompt that is meant to induce the `persona' of an experienced resume writer into the LLM, followed by the task prompt to extract the important job details. More specifically, for the system prompt, we describe a persona specifically that mimics the tone, style, and focus of professional resumes, such as: ``You are a seasoned career advising expert in crafting resumes and cover letters, boasting a rich 15-year history dedicated to mastering this skill...". We empirically see that priming the LLM via this prompt greatly improves the performance of the resulting system, since it enables the outputs to adhere to the context and task-specific conventions and language. 

Then the task prompt is designed to guide the LLM to extract the important job details from the provided job description and output these as a structured JSON. By combining these prompts, the LLM accurately extracts and structures key information, presented in a structured JSON format. Accurately extracting this data is crucial for the next step of the pipeline - that is the resume tailoring step. This extracted data typically includes essential elements like title, keywords, purpose, responsibilities, qualifications (both required and preferred), company name, and company information. We refer to this JSON of job details as \textbf{JobDetails} in the remainder of this paper.

\subsection{Resume Generator}

This is the main component of our framework, where the processing of the resume takes place and thereby a modified version of the user's resume is generated after being tailored towards the specific job. The two inputs to this component are the \textbf{UserData} and the \textbf{JobDetails} from the previous two steps. Now, the \textbf{UserData} is processed section by section: say a typical resume has the following sections: \{Personal Details, Education, Work Experience, Skills, Achievements, Projects\}, etc. We choose to process the resume by dividing it into sections since it follows the natural structure of how a standard resume is formatted and organized. Additionally, LLMs have restrictions on the context length they can handle, and feeding the entire resume text at once would exceed the context window. Furthermore, LLMs have been shown to struggle to extract information from very long contexts, and often overlook information that is in the middle of a long context~\cite{liu2023lost}. Therefore processing each section at a time alleviates all these issues. We iterate over each of the \textbf{UserData} sections as follows:
first, the personal details section is extracted without change, since we do not want the LLM to change any factual information that is in this section, such as phone number, address, or name.
Next, we loop over the sections in \textbf{UserData}, prompt the chosen LLM with a section-specific prompt along with the \textbf{UserData} section data and the \textbf{JobDetails}. For each section, the LLM is tasked with re-creating the section after analyzing the user's resume and retaining, emphasizing, or removing points to make that section more aligned with the specific job. At the end of processing each of these sections, we collect the response of the LLM. 
Finally, these are combined into a JSON.

The system prompt we use here contains instructions on how to craft the resume based on the job description in \textbf{JobDetails}. To do this we use best practices as highlighted by career counselors and professional resume writers as in ~\cite{linkedin_1,Res_2}. More specifically, we carefully instruct the LLM to follow these principles:
\begin{enumerate}
\item \textit{Focus}: Craft relevant achievements aligned with the job description. 
\item \textit{Honesty}: Prioritize truthfulness and objective language.
\item \textit{Specificity}: Prioritize relevance to the specific job over general achievements.
\item \textit{Style}: 
\begin{enumerate}
    \item \textit{Voice}: Use active voice whenever possible. 
  \item \textit{Proofreading}: Ensure impeccable spelling and grammar.
\end{enumerate}
\end{enumerate}

These carefully tailored instruction-style prompts leverage the instruction-understanding capabilities of these state-of-the-art LLMs and strive to emulate the skill and style of professional resume writers who are experienced in this task.

Our tool provides the user the option to choose which LLM to use in the pipeline. Currently, we have support for OpenAI models (default is \texttt{gpt-4-1106-preview}), and Google Deepmind's Gemini~\cite{team2023gemini} \footnote{https://deepmind.google/technologies/gemini/} (default version is \texttt{gemini-pro}). For the \texttt{gpt-4} models we use the ChatCompletion backend and we use the \texttt{json\_object} output format in order to easily handle LLM outputs programmatically, by simply casting these JSON outputs into a Python dictionary. For \texttt{gemini}, we use a parsing function we defined to parse the generated output into a JSON format. Finally, we use a \LaTeX engine to convert this JSON to a visually appealing and professionally formatted PDF document, allowing for customization based on various resume templates. Alongside this resume generation, our tool also gives the user the option to generate a cover letter that is aligned with the job as well.

\begin{figure}
    \centering
    \includegraphics[width=0.8\columnwidth]{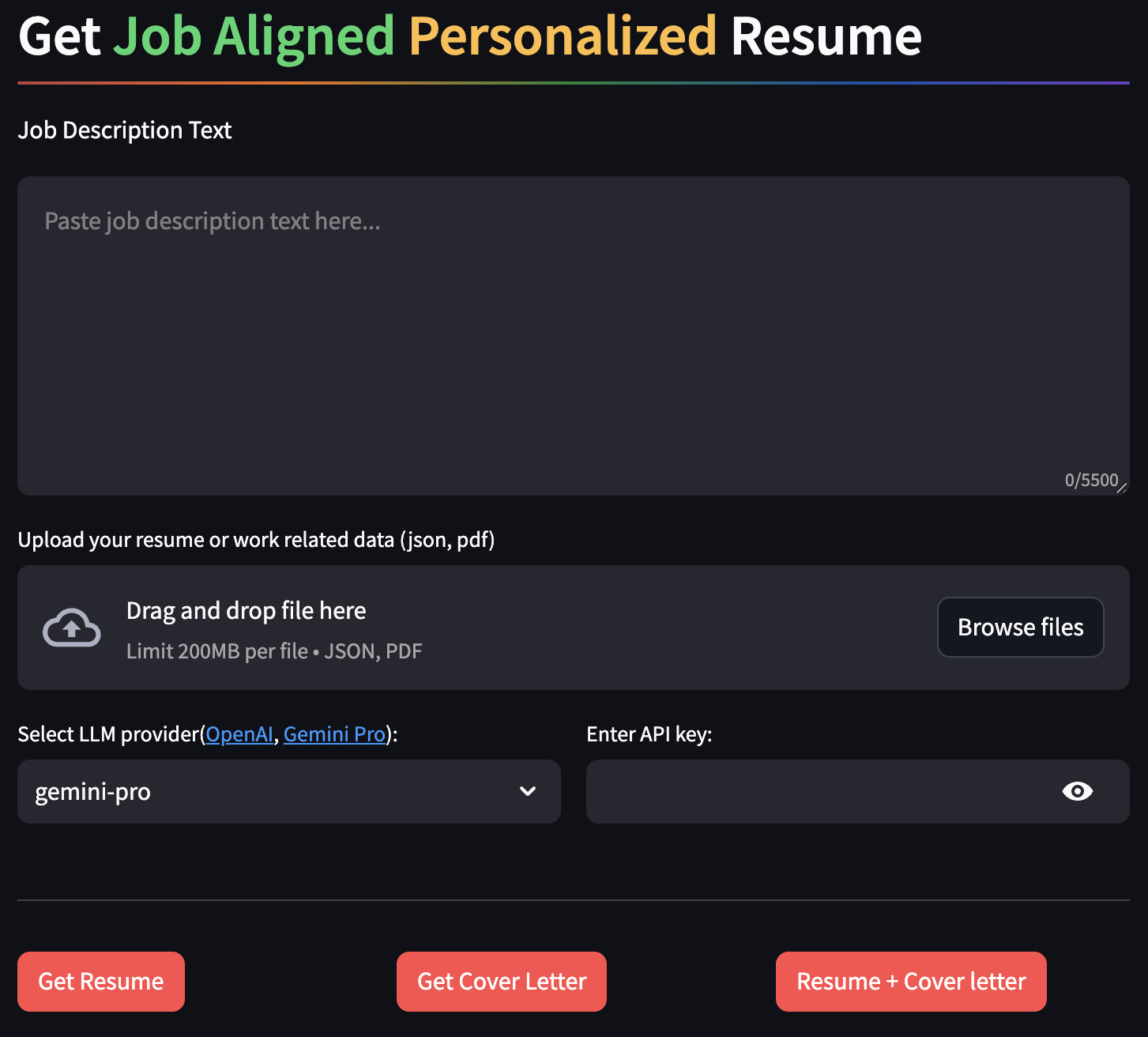}
    \caption{Our easy-to-use User Interface.}
    \label{fig:ui}
\end{figure}

\begin{figure}
    \centering
    \includegraphics[width=0.8\columnwidth]{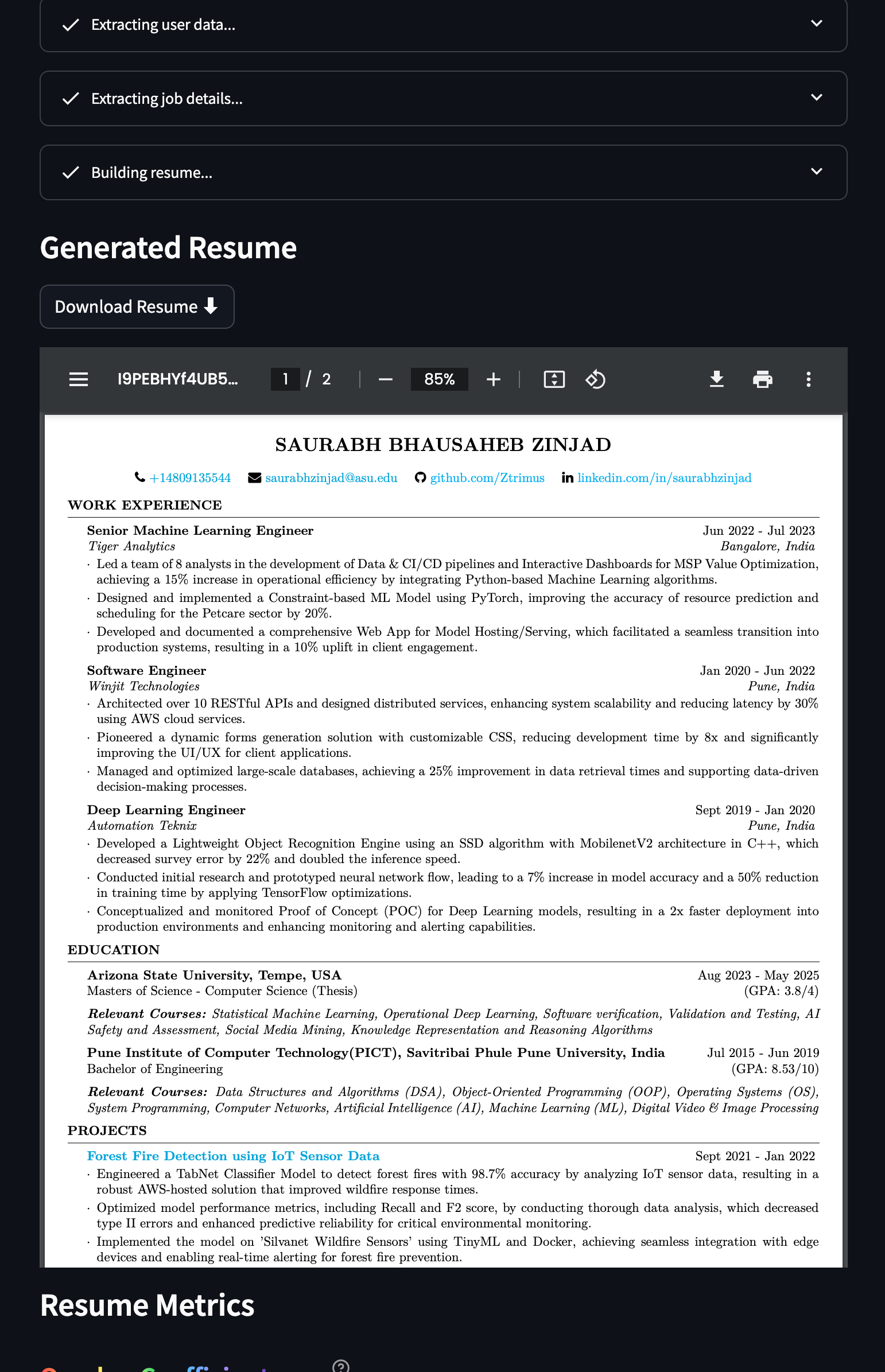}
    \caption{Truncated screenshot of \textsc{ResumeFlow} generating a job-aligned resume.}
    \label{fig:screen}
\end{figure}

\section{Evaluation}

In this section, we describe how we evaluate our system on the basis of \textit{goodness} of the generated resumes. Unlike standard natural language processing tasks that have well-defined benchmarks and evaluation metrics, automated evaluation of such a system solving this novel task is quite challenging. However, we introduce two metrics to evaluate generated resumes on two dimensions: \textit{job\_alignment} and \textit{content\_preservation}. We formulate these metrics as:

\textbf{\textit{job\_alignment}}: This metric measures the alignment between the generated resume and the job details. Ideally, we would want a higher score since this would mean the resume generated by our tool is well-tailored to the specific job the user is applying to. We measure this in both the token space and the latent space. In the token space, we define $job\_alignment_{token}$ as the overlap coefficient ~\cite{vijaymeena2016survey} between the unique words in the generated resume and the unique words in the job description. In the latent space, we define $job\_alignment_{latent}$ as the cosine similarity between vector embeddings of the generated resume and the job description text. We use Gemini and OpenAI embeddings\footnote{\texttt{models$\slash$embedding-001} for Gemini, \texttt{text-embedding-ada-002} for OpenAI.} to embed both the texts. We formulate these as:

\begin{equation}
    job\_alignment_{token} = \frac{|\mathcal{W}(\mathcal{D}^{gen}) \cap \mathcal{W}(\mathcal{J})|}{min(|\mathcal{W}(\mathcal{D}^{gen})|, |\mathcal{W}(\mathcal{J})|)}
\end{equation}

where $\mathcal{D}^{gen}$ refers to the resume generated by our tool, $\mathcal{J}$ refers to the job description, $\mathcal{W}(x)$ refers to the set of unique words in $x$.

\begin{equation*}
    job\_alignment_{latent} = cosine\_sim(\mathcal{E}(\mathcal{D}^{gen}), \mathcal{E}(\mathcal{J}))
\end{equation*}

where $cosine\_sim(\cdot,\cdot))$ refers to the cosine similarity, $\mathcal{E}(\cdot)$ refers to the encoding used to embed the text.

\textbf{\textit{content-preservation}}: Since hallucination is a known issue even in state-of-the-art LLMs, we would want to verify how well our \textsc{ResumeFlow} pipeline is able to preserve content between the user-provided resume and the generated one. Ideally, we would want high values for this, since low values of content preservation might imply hallucination by the LLM.  Similar to the \textit{job\_alignment} metric, we measure in both the token space and the latent space. In the token space, we formulate this as the overlap coefficient between the user's original resume and the \textsc{ResumeFlow} generated resume:

\begin{equation*}
    content\_preservation_{token} = \frac{|\mathcal{W}(\mathcal{D}^{gen}) \cap \mathcal{W}(\mathcal{D}^{user})|}{min(|\mathcal{W}(\mathcal{D}^{gen})|, |\mathcal{W}(\mathcal{D}^{user})|)}
\end{equation*}

where  $\mathcal{D}^{user}$ refers to the original resume input by the user to the \textsc{ResumeFlow} tool.
Similarly, in the latent space, we consider the cosine similarity between the generated resume and the original resume:

\begin{equation*}
    content\_preservation_{latent} = cosine\_sim(\mathcal{E}(\mathcal{D}^{gen}), \mathcal{E}(\mathcal{D}^{user}))
\end{equation*}

Note that this metric is loosely inspired by summarization metrics such as the ROUGE score commonly used in NLP~\cite{lin2004rouge} and also bears similarity with recently proposed hallucination detection metrics~\cite{forbes2023metric} such as AlignScore~\cite{zha2023alignscore}.

The combination of these two metrics: \textit{job\_alignment} and \textit{content\_preservation} provide an idea of how good the generated resume is. Low values of \textit{content-preservation} coupled with high values of \textit{job-alignment} is particularly undesirable since it implies that the LLM has possibly hallucinated major parts of the generated resume in order to make it very suitable for the job description. This particular case might be considered unethical since it may seem dishonest from the user's side. We report these scores on the user interface as well, so the end user gets an idea of how well the generated resume is, before directly using it.

\section{Conclusion \& Future Work}

In this work, we introduce an easy-to-use pipeline \textsc{ResumeFlow} to help a job applicant tailor their resume to a specific job posting by simply providing their resume and the corresponding job description to our LLM-facilitated pipeline. While our tool currently supports two state-of-the-art models: GPT-4 and Gemini, future improvements to the tool could incorporate more LLMs, preferably open-source ones. With any LLM-powered system, the users of the system need to be aware of the phenomenon of hallucination and how to mitigate instances of hallucinations. To that end, future work may utilize emerging concepts such as retrieval augmented generation and generation using knowledge graphs.

\bibliographystyle{ACM-Reference-Format}
\balance
\bibliography{sample-base}

\appendix

\end{document}